\documentclass[p1]{nature}
\usepackage{graphicx,epsfig,dsfont,amssymb,amsmath,amsthm,amsfonts,amsbsy,mathrsfs,amscd,epstopdf,multirow,adjustbox}
\usepackage{float}
\makeatletter
\let\saved@includegraphics\includegraphics
\AtBeginDocument{\let\includegraphics\saved@includegraphics}
\renewenvironment*{figure}{\@float{figure}}{\end@float}
\makeatother
\usepackage[tight]{subfigure}
\usepackage{subfloat}
\usepackage{url}
\usepackage{threeparttable}
\usepackage{booktabs}
\usepackage[T1]{fontenc}
\usepackage{xspace}
\usepackage{color}
\usepackage{rotating}
\usepackage[pagewise]{lineno}
\title{CAMEL2: Enhancing weakly supervised learning for histopathology images by incorporating the significance ratio}

\author{Gang Xu$^{1,2,3,\dagger}$, Shuhao Wang$^{4,\dagger,*}$, Lingyu Zhao$^{5,\dagger}$, Xiao Chen$^{6}$, Tongwei Wang$^{7}$, Lang Wang$^{4}$, Zhenwei Luo$^{1,2,3}$, Dahan Wang$^{8}$, Zewen Zhang$^{4}$, Aijun Liu$^{9}$, Wei Ba$^{10}$, Zhigang Song$^{10}$, Huaiyin Shi$^{10}$, Dingrong Zhong$^{5,*}$, Jianpeng Ma$^{1,2,3,*}$}

\begin{document}

\maketitle

\begin{affiliations}
\item Multiscale Research Institute of Complex Systems, Fudan University, Shanghai, China
\item Zhangjiang Fudan International Innovation Center, Fudan University, Shanghai, China
\item Shanghai AI Laboratory, Shanghai, China
\item Thorough Lab, Thorough Future, Beijing, China
\item Department of Pathology, China-Japan Friendship Hospital, Beijing, China
\item College of Mathematics and Data Science (Software College), Minjiang University, Fuzhou, China
\item College of Future Technology, Peking University, Beijing, China
\item Fujian Key Laboratory of Pattern Recognition and Image Understanding, School of Computer and Information Engineering, Xiamen University of Technology, Xiamen, China
\item Department of Pathology, Seventh Medical Center, Chinese PLA General Hospital, Beijing, China
\item Department of Pathology, First Medical Center, Chinese PLA General Hospital, Beijing, China

\thanks{$^\dagger$ These authors contributed equally to this work.}

\thanks{$^*$ Corresponding authors.}

\end{affiliations}

\begin{abstract}

Histopathology image analysis plays a crucial role in cancer diagnosis. However, training a clinically applicable segmentation algorithm requires pathologists to engage in labour-intensive labelling. In contrast, weakly supervised learning methods, which only require coarse-grained labels at the image level, can significantly reduce the labeling efforts. Unfortunately, while these methods perform reasonably well in slide-level prediction, their ability to locate cancerous regions, which is essential for many clinical applications, remains unsatisfactory. Previously, we proposed CAMEL, which achieves comparable results to those of fully supervised baselines in pixel-level segmentation. However, CAMEL requires 1,280$\times$1,280 image-level binary annotations for positive WSIs. Here, we present CAMEL2, by introducing a threshold of the cancerous ratio for positive bags, it allows us to better utilize the information, consequently enabling us to scale up the image-level setting from 1,280$\times$1,280 to 5,120$\times$5,120 while maintaining the accuracy. Our results with various datasets, demonstrate that CAMEL2, with the help of 5,120$\times$5,120 image-level binary annotations, which are easy to annotate, achieves comparable performance to that of a fully supervised baseline in both instance- and slide-level classifications.

\end{abstract}

\section*{Introduction}

Histopathology image analysis is considered the gold standard for cancer diagnosis. However, developing a clinically applicable histopathological diagnosis system for cancerous region segmentation requires extensive and time-consuming labelling processes at the pixel level that must be performed by experienced pathologists \cite{RN1}. A previous study \cite{RN2} showed that a pathologist requires approximately 15 minutes to half an hour to examine a gigapixel whole-slide image (WSI). Furthermore, due to the potential for oversight and mislabelling of cancerous regions, examinations by three or more pathologists per WSI are often necessary. Unfortunately, experienced pathologists with years of training are scarce, and their time is extremely valuable given the high patient volume in hospitals. To address this challenge, various weakly supervised methods that require less labelling have been proposed \cite{RN3,RN4,RN5,RN6,RN7,RN8,RN9,RN10,RN11}.

Histopathology images, in comparison to natural images \cite{RN12,RN13,RN14} or other medical images like CT scans \cite{RN15} or MRI images \cite{RN16}, are typically characterized by their large size, ranging from 100 million to 10 billion pixels. Consequently, directly transferring methods between these fields is almost impractical \cite{RN10}. Moreover, histopathology images exhibit complicated cell morphologies and tissue structures \cite{RN17} . Therefore, methods based on seed growing \cite{RN18} or adversarial erasing \cite{RN19} may not be suitable. In this context, multiple instance learning (MIL) has emerged as a general solution for weakly supervised learning approaches in histopathology image analysis. MIL involves splitting an image into instances and considering instances from the same image as belonging to the same bag. During training, annotations are only available at the bag level.

In recent years, some weakly supervised methods \cite{RN3,RN6,RN7,RN10} that only require whole slide image-level (WSI-level) annotations have been proposed, largely alleviating the labelling burden. Most of these methods are based on MIL. First, they select a subset of patches from each WSI and utilize models pretrained with ImageNet \cite{RN20} to generate patch-level features. Then, the patch-level features are aggregated into slide-level representations using various algorithms. One commonly used method in the field is CLAM \cite{RN6}, which segments each slide into smaller patches (256$\times$256) and employs a pretrained CNN model \cite{RN20} for feature extraction. Finally, an attention-based pooling function is utilized to aggregate patch-level features into slide-level predictions. Another method, TransMIL \cite{RN7}, uses a self-attention-based aggregation method derived from a transformer model \cite{RN21} to represent slide-level predictions. DTFD-MIL \cite{RN10} introduces Grad-CAM \cite{RN22} into an attention-based aggregation method to perform double-tier feature distillation, resulting in improved slide-level representations. Although these methods, which solely use slide-level annotations, achieve reasonable performance in slide-level classification, their ability to locate cancerous regions is often unsatisfactory, which is crucial for many clinical applications. This limitation may arise from insufficient slide-level annotations during the training process. For instance, the CAMELYON16 dataset \cite{RN23}, one of the largest public histopathology WSI datasets, only contains 399 haematoxylin-eosin (H\&E)-stained WSIs of lymph node sections. Among them, the official training set includes only 240 WSIs (110 positives, 130 negatives), meaning that only 240 slide-level annotations are available for training the aforementioned algorithms.

To achieve comparable performance in locating cancerous regions using fully supervised learning methods, the introduction of additional and easily obtained annotations may be helpful. For example, CDWS-MIL \cite{RN11} incorporates cancerous area constraints into its framework to achieve improved results. However, during its training process, a large number of 500$\times$500 image-level labels (positive or negative) and their corresponding cancer ratios are needed, limiting its practicality. Wang et al. utilized LC-MIL \cite{RN9} to refine roughly drawn coarse annotations on WSIs. In our previous work, CAMEL \cite{RN8}, we included additional 1,280$\times$1,280 image-level labels (positive or negative) for each slide and considered each image as a bag under the MIL setting, ultimately achieving comparable performance in terms of instance-level classification and pixel-level segmentation with that of a fully supervised baseline (FSB). However, in addition to the 240 slide-level labels in CAMELYON16, CAMEL requires an additional 5,056 positive 1,280x1,280 image-level labels (20$\times$ magnification) for training, which is both costly and clinically infeasible.

In this research, our objective is to train an instance-level (320$\times$320) classifier for locating cancerous regions in a weakly supervised manner while achieving performance comparable to that of a FSB, thus greatly improving the interpretability of weakly supervised learning, which is crucial for many clinical applications. To this end, we propose CAMEL2, which leverages a straightforward strategy by setting a positive threshold ($t$\%), i.e., the significance ratio, for positive bags to better utilize the information from each WSI. As a result, it successfully scales up the image-level setting from 1,280$\times$1,280 in CAMEL to 5,120$\times$5,120 in this study while achieving comparable performance in terms of both instance- and slide-level classifications to that of a FSB. Furthermore, by integrating CAMEL2 with other models, it can be tailored for tasks that exclusively rely on slide-level annotations yet demand relatively high interpretability. For example, it can be used to identify regions associated with ALK rearrangement in lung adenocarcinomas.

\begin{figure}
    \centering
    \includegraphics[width=0.8\textwidth]{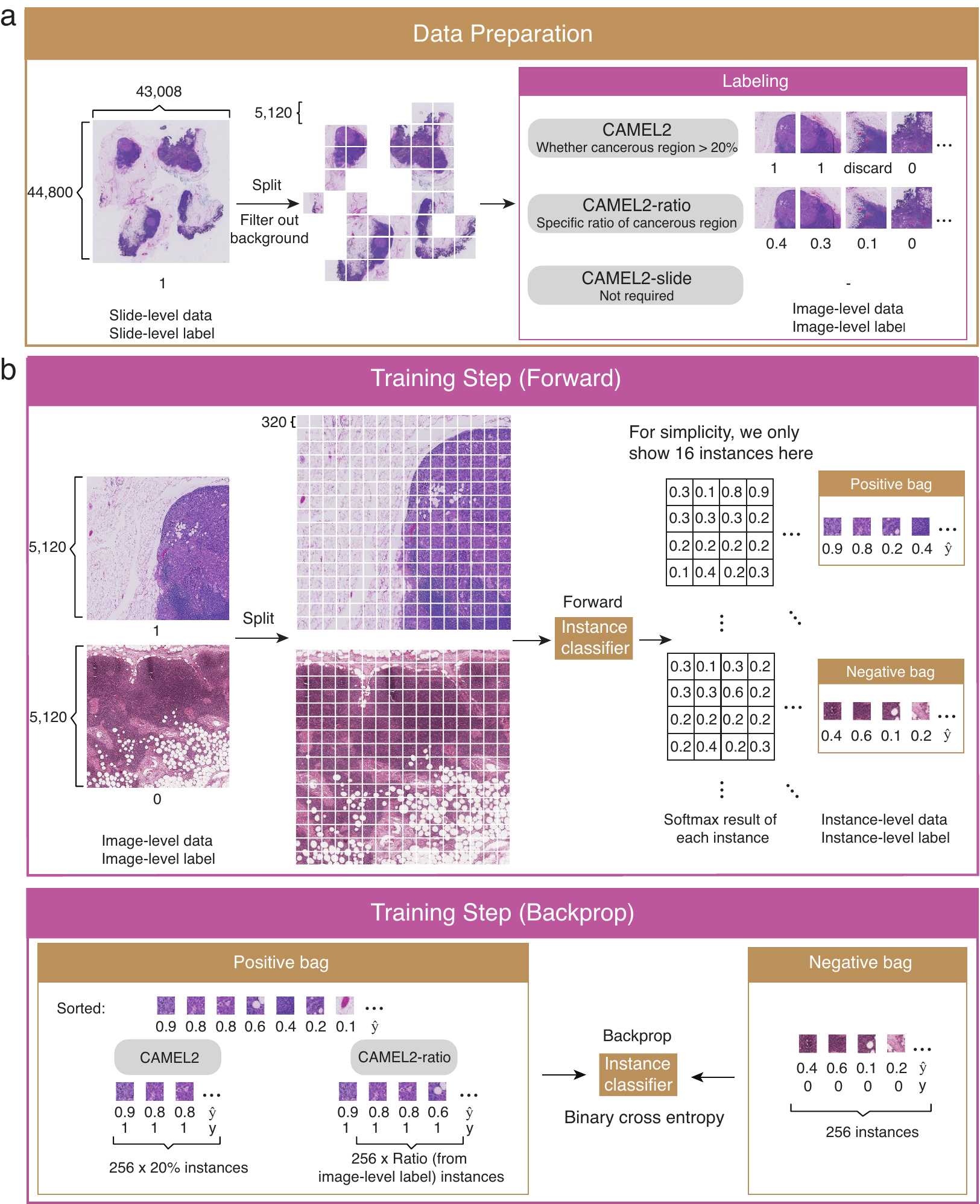}
    \caption{{\bf The framework of CAMEL2.} {\bf a}, Flowchart of data preparation for different variants of CAMEL2. Here, we use a WSI with 44,800$\times$43,008 pixels under 20$\times$ magnification in CAMELYON16 as an example. After splitting into 5,120$\times$5,120 image-level data and removing background, only 33 images remain for manual labelling. {\bf b}, Training procedure of CAMEL2. 5,120$\times$5,120 image-level data are considered as bags and divided into 256 320$\times$320 instances. The instances with the top $t$\% highest positive probability for positive WSIs are selected and labelled positive. In CAMEL2 and CAMEL2-ratio, the thresholds for positive bags are set to 20\% and the ratio constraint specific to each bag, respectively.}
    \label{fig:framework}
\end{figure}

\section*{Results}
{\bf Overview.}
Figure \ref{fig:framework} illustrates the three variants of CAMEL2 tested in this study. These variants employ different types of annotations. Similar to CLAM \cite{RN6}, CAMEL-slide is the least labour-intensive method, requiring only slide-level (WSI-level) annotations. Conversely, CAMEL2-ratio is the most labour-intensive method, necessitating additional 5,120x5,120 image-level annotations specifying a specific cancerous region ratio for each positive WSI. For CAMEL2, in addition to slide-level annotations, we need pathologists to annotate the 5,120$\times$5,120 images from positive WSIs as positive (cancerous ratio $\ge$ 20\%) or negative (cancer free), and discard the images with cancerous regions with ratios less than 20\%. In this study, we evaluate CLAM, CAMEL, and three CAMEL2 variants using the WSIs at 20$\times$ magnification.

In contrast to using professional tools to meticulously outline cancerous areas, our approach of employing binary annotations for a limited set of images in CAMEL2 considerably eases the workload for pathologists and reduces the technical demands. Furthermore, we only classify an image as positive if over 20\% of its area is cancerous. This threshold significantly simplifies the task of identifying minor cancerous regions. In Table \ref{tab:s1r}, we have detailed the average time spent by three pathologists annotating 5 positive WSIs from the CAMELYON16 dataset, utilizing the iPad-based annotation system developed by Thorough Future \cite{RN1}.

\begin{figure}
    \centering
    \includegraphics[width=0.9\textwidth]{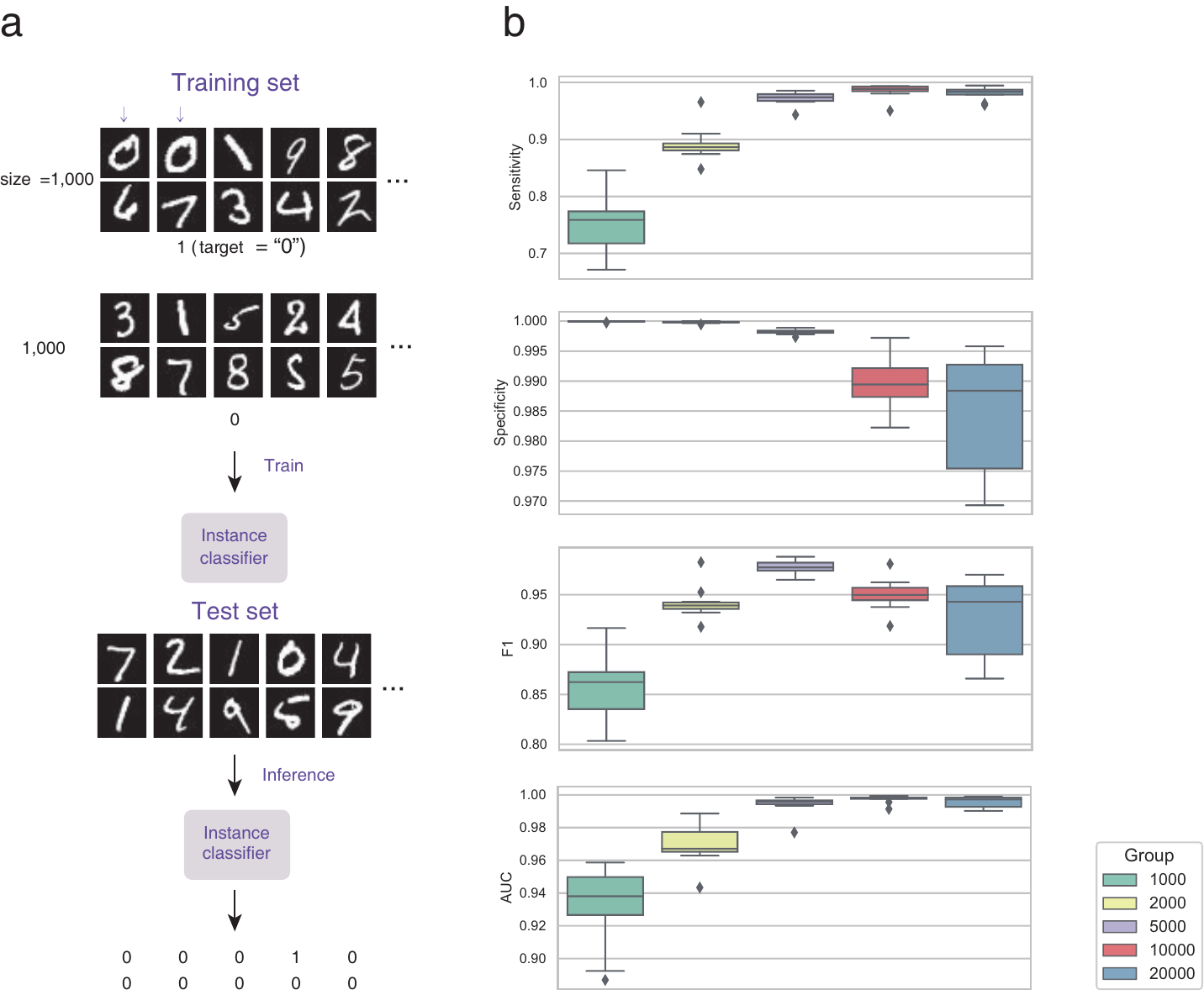}
    \caption{{\bf Multiple instance MNIST task.} {\bf a}, Illustration of a binary classification experiment with the MNIST dataset. In this case, each bag contains 1,000 instances. A bag is labelled as positive if it contains at least one instance with a handwriting label of "0"; otherwise, it is labelled as negative. {\bf b}, Binary classification results of CAMEL2 with the MNIST test set. Each group represents experiments with sizes of 1,000, 2,000, 5,000, 10,000, and 20,000, respectively. Ten experiments are conducted with the target label varying from "0" to "9". Since the number of positive instances in each positive bag is randomly sampled from a uniform distribution U(0, 1,000), the proportion of target instances in positive bags for each group ranges from 0-100\%, 0-50\%, 0-20\%, 0-10\%, and 0-5\%, respectively.}
    \label{fig:mnist}
\end{figure}

{\bf Multiple instance MNIST.}
To evaluate the framework of CAMEL2, we conduct experiments using the MNIST dataset \cite{RN24}. In these experiments, we treat the individual samples in MNIST as instances and randomly group them into bags. Two hyperparameters are used: \emph{size} determines the number of instances in each bag, and \emph{target} represents a specific handwriting label ("0-9"). If the label of at least one instance is the target, the bag is labelled as positive; otherwise, it is labelled as negative. Figure \ref{fig:mnist}a provides an illustration of an experiment where the size is set to 1000 and the target is set to "0".

We conduct 5 groups of experiments, varying the size as follows: 1,000, 2,000, 5,000, 10,000, and 20,000 instances per bag. We also change the target label from "0" to "9," one label at a time, resulting in a total of 10 experiments. During training, we divide the instances in the training set into negative-instance and positive-instance groups based on the target label. All of the instances in the negative bag are randomly sampled from the negative-instance group. The number of positive instances in each positive bag is derived from a uniform distribution U(0, 1,000), and the positive and negative instances are proportionally and randomly sampled from the positive-instance and negative-instance groups, respectively. Consequently, for positive bags, the proportion of target instances varies across different ranges: 0-100\%, 0-50\%, 0-20\%, 0-10\%, and 0-5\%. All of them are annotated as positive, independent of the threshold percentage ($t$\%) used in CAMEL2. In contrast, negative bags contain no target instances. Throughout all experiments, we set the threshold ($t$\%) for CAMEL2 during the training process at 10\%.

\begin{figure}
    \centering
    \includegraphics[width=0.9\textwidth]{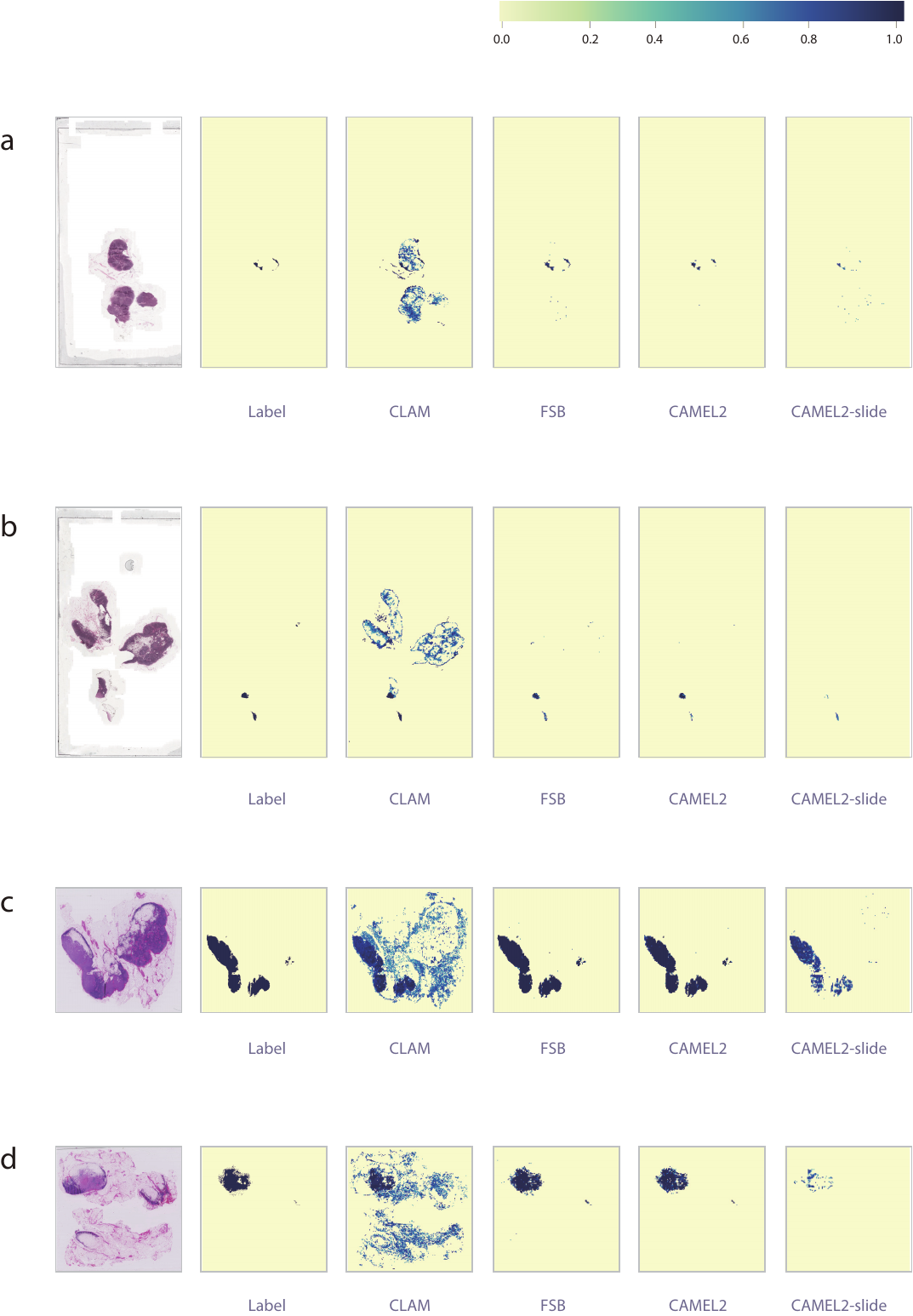}
    \caption{{\bf Performance of different methods using CAMELYON16 test set WSIs with a relatively large ratio of cancerous regions.}}
    \label{fig:large}
\end{figure}

As shown in Figure \ref{fig:mnist}b, as the proportion of target instances in positive bags decreases, the sensitivity of CAMEL2 increases, while the specificity decreases. For two extreme cases, if we set the size to 1,000 (i.e., the proportion of target is 0-100\%), the threshold of CAMEL2 (10\%) is much lower than the actual ratio of target instances, potentially leading to inadequate learning for target instances. On the other hand, if we set the size to 20,000 (i.e., the proportion of target is 0-5\%), the higher threshold of CAMEL2 could result in a false-positive issue. In terms of F1 and the AUC, CAMEL2 demonstrates satisfactory identification of target instances in a weakly supervised manner, regardless of the proportion of target instances in the bags. The results also suggest that setting the threshold of CAMEL2 more precisely based on prior domain knowledge could enhance the overall performance.

{\bf CAMELYON16.}
In Supplementary Table \ref{tab:s1}, we list the 320$\times$320 patch-level classification results of different methods with the official test set of CAMELYON16. Each model employs the same ResNet-18 \cite{RN25} backbone for fair comparison. The results demonstrate that when using the 1,280$\times$1,280 image-level annotations with specific cancerous ratio annotations, CAMEL2-ratio achieves comparable performance to that of the FSB. By maximizing the utilization of positive instances in each positive bag based on the ratio of cancerous regions, CAMEL2-ratio outperforms the traditional MIL method, which only considers one instance per positive bag.

\begin{figure}
    \centering
    \includegraphics[width=0.9\textwidth]{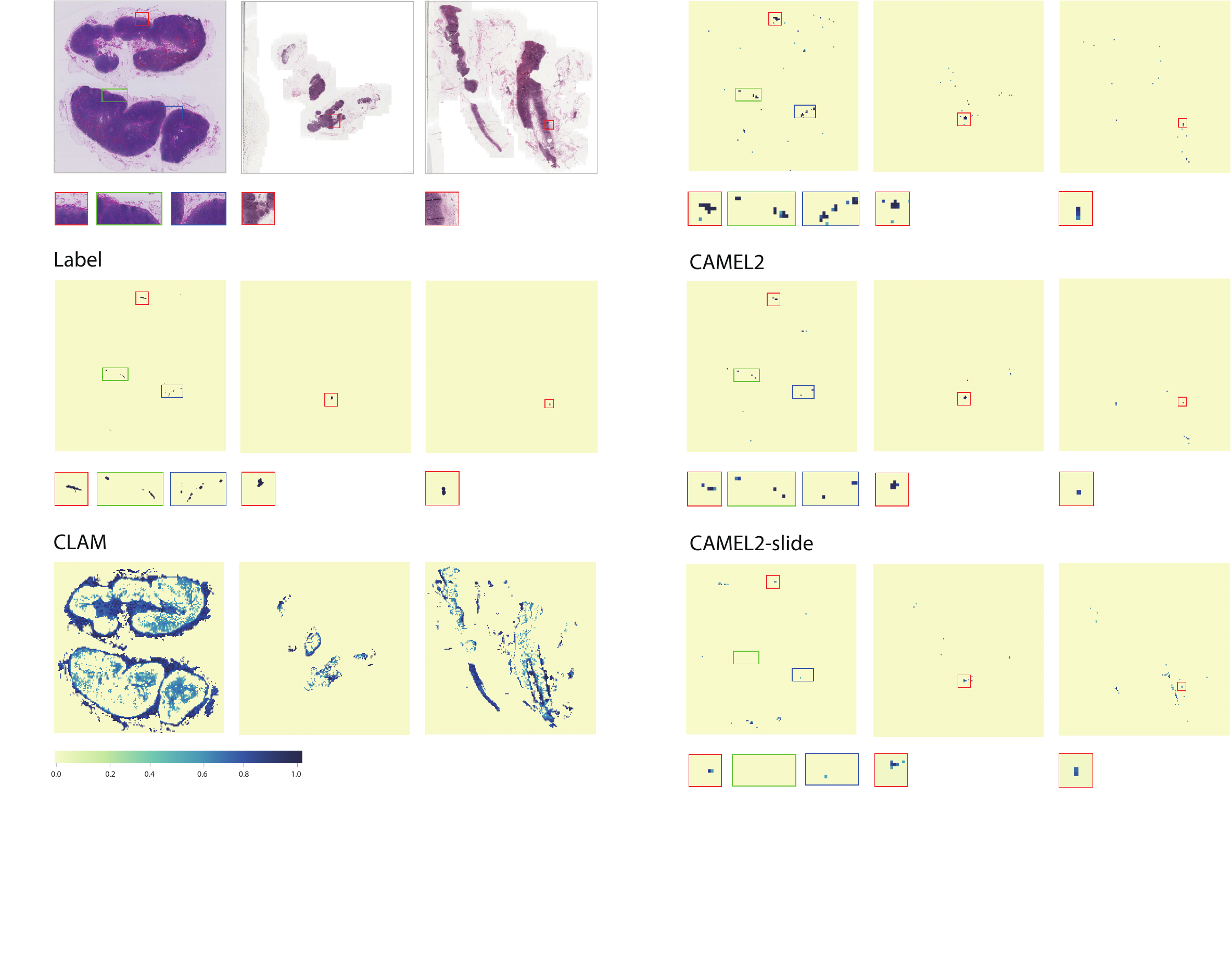}
    \caption{{\bf Performance of different methods using CAMELYON16 test set WSIs with a relatively small ratio of cancerous regions.}}
    \label{fig:small}
\end{figure}

When the size of image-level annotations is set to 5,120$\times$5,120, a 16$\times$ reduction in annotations causes a decrease in the performance of each method. However, CAMEL2 still achieves comparable performance to that of the FSB in terms of F1 and AUC. Supplementary Figure \ref{fig:lsen} highlights some cases where CAMEL2 exhibits low sensitivity with the test set of CAMELYON16. The relatively low sensitivity is primarily due to the insufficient identification of large cancerous regions, which is somewhat acceptable in this task. In such cases, CAMEL2 generally locates the cancerous regions but does not fully identify all of them. Complete results of CAMEL2 and the FSB for all 129 WSIs in the official test set of CAMELYON16.

In addition, CAMEL2, which is trained using 322 positive images with binary annotations, achieves performance comparable to that of CAMEL2-ratio, which requires annotations of specific ratios of cancerous regions for 727 positive images and thus may significantly reduce the labelling effort. The results also indicate that the retraining step inspired by CAMEL \cite{RN8} improves the performance of both CAMEL2-ratio and CAMEL2 in terms of F1 and the AUC. Furthermore, when using 156 images split from positive WSIs with a cancerous ratio > 50\% and setting the threshold of CAMEL2 to 50\% during training, the performance of CAMEL2 ($t$ = 50\%) decreases, which is likely due to the reduced number of positive image-level samples.

Label ambiguity can occur when pathologists are uncertain whether the cancerous area is above or below the 20\% threshold. To simulate this situation, we annotate images as positive at a 20\% threshold, while employing 25\% and 35\% thresholds during the training phases of CAMEL2 (referenced as CAMEL2 ($wt$ = 25\%) and CAMEL2 ($wt$ = 35\%) in Supplementary Table \ref{tab:s1}). The results indicate that higher training thresholds can increase the rate of false positives, enhance sensitivity, but lower specificity. Identifying the optimal threshold ($t$\%) automatically for specific datasets remains an important issue, which we aim to address in our future work.

In Figure \ref{fig:large}, we present the predictions of some positive WSIs from the CAMELYON16 test set, which have a relatively large ratio of the cancerous region. The results demonstrate that both CAMEL2 and CAMEL2-slide effectively locate most of the cancerous regions and offer better interpretability compared to CLAM \cite{RN6}. CLAM relies on models pretrained with ImageNet \cite{RN20} for patch-level feature extraction and primarily focuses on developing advanced aggregation algorithms for slide-level prediction.

Figure \ref{fig:small} showcases the predictions of some challenging targets that have a relatively small ratio of cancerous region. Our models are capable of delivering reasonably accurate predictions. However, due to the limited number of slide-level annotations, the performance of CAMEL2-slide is inferior to that of CAMEL2 (see Figure \ref{fig:small}a).

\begin{figure}
    \centering
    \includegraphics[width=0.9\textwidth]{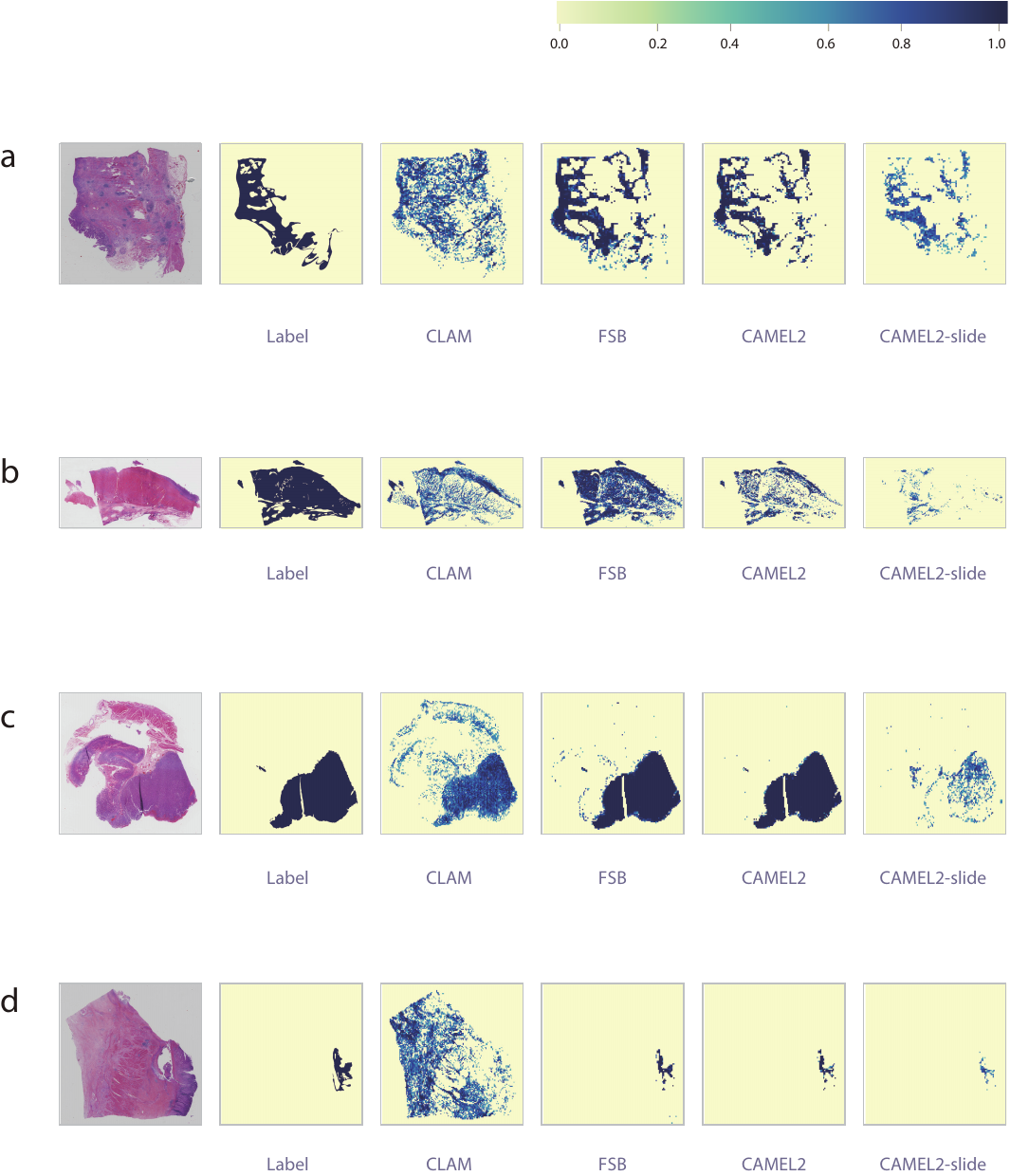}
    \caption{{\bf Performance of different methods using the test set WSIs of the gastric cancer dataset.}}
    \label{fig:gastric}
\end{figure}

{\bf In-house gastric cancer dataset.}
We assign the malignant label to both high-grade intraepithelial neoplasia and carcinoma because both lesions require surgical intervention. In Supplementary Figure \ref{fig:roc_gastric}, we show the ROC curve of the 320$\times$320 patch-level classification results of different methods using the test set of gastric cancer dataset. Detailed comparisons are listed in Supplementary Table \ref{tab:s2}. Our results indicate that with the help of 5,120$\times$5,120 image-level binary annotations, CAMEL2 achieves comparable performance to that of the FSB in terms of F1 and AUC. The cancerous regions predicted for the gastric cancer dataset are shown in Figure \ref{fig:gastric}b.

{\bf In-house cervical cancer dataset.}
Similar to the in-house gastric cancer dataset, the malignant label includes high-grade intraepithelial neoplasia and carcinoma. Supplementary Figure \ref{fig:roc_cervix} presents the ROC curve of the 320$\times$320 patch-level classification results of different methods using the test set of our collected histopathology images of cervical cancer. Detailed comparisons are listed in Supplementary Table \ref{tab:s3}. Precisely identifying cancerous regions for cervical cancer is challenging, even for the FSB, which may be caused by the diversities of cell morphology and complicated tissue structure (see Supplementary Figure \ref{fig:cervix_fp}). Our results demonstrate that CAMEL2 also achieves comparable performance to that of the FSB.

\begin{figure}
    \centering
    \includegraphics[width=0.9\textwidth]{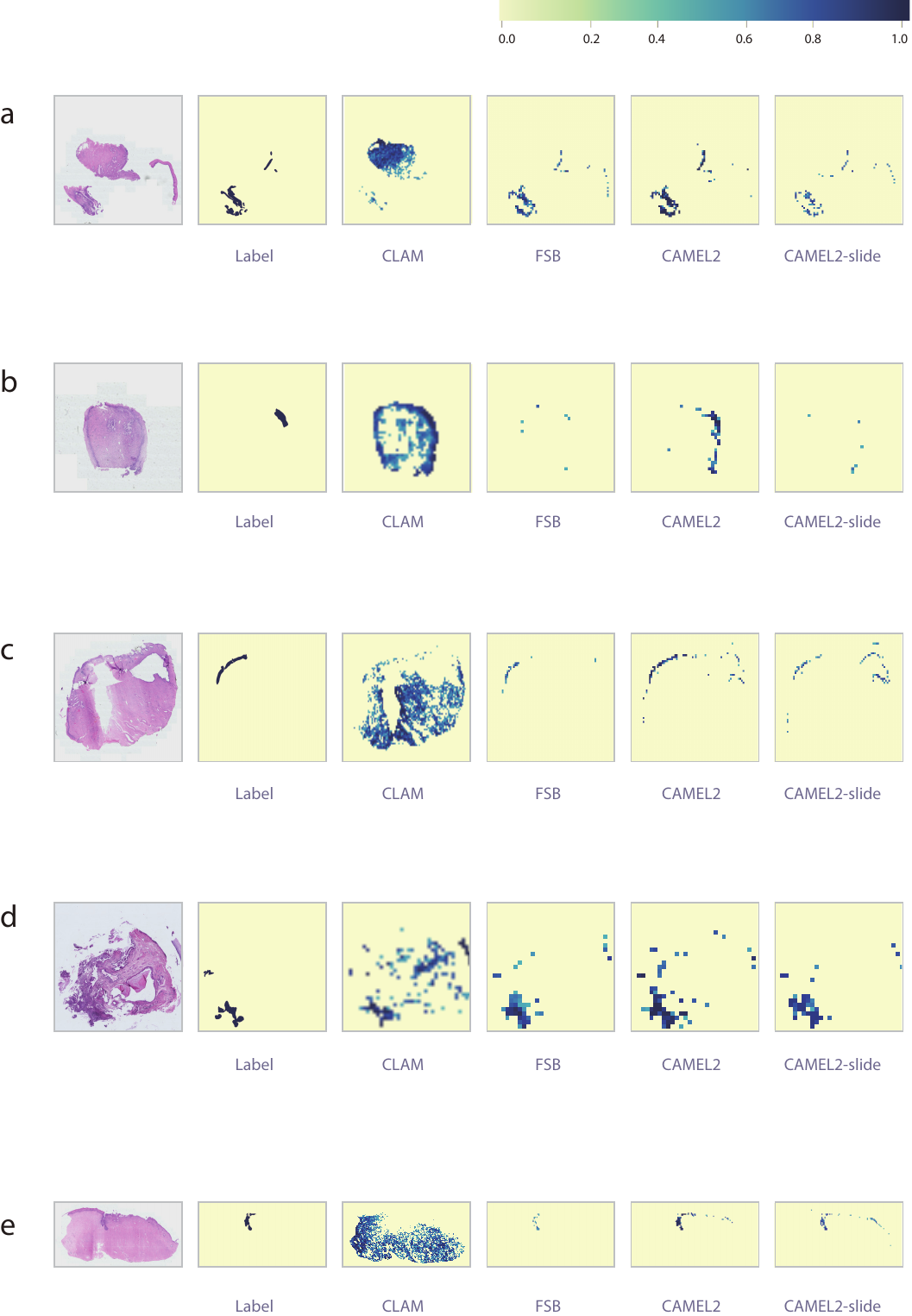}
    \caption{{\bf Performance of different methods using test set WSIs of the cervical cancer dataset with a relatively large ratio of cancerous regions.}}
    \label{fig:cervix}
\end{figure}

In addition, we show the predictions of some positive WSIs from the test set that have relatively large and small ratios of cancerous regions in Figure \ref{fig:cervix} and Supplementary Figure \ref{fig:cervix_region}, respectively. The results indicate that both CAMEL2 and CAMEL2-slide effectively identify most of the cancerous regions. However, due to the currently limited data and annotations, certain cancerous and noncancerous regions are indistinguishable, which is a challenge even for the FSB. Moreover, our models demonstrate better interpretability than that of CLAM.

\begin{figure}
    \centering
    \includegraphics[width=1.0\textwidth]{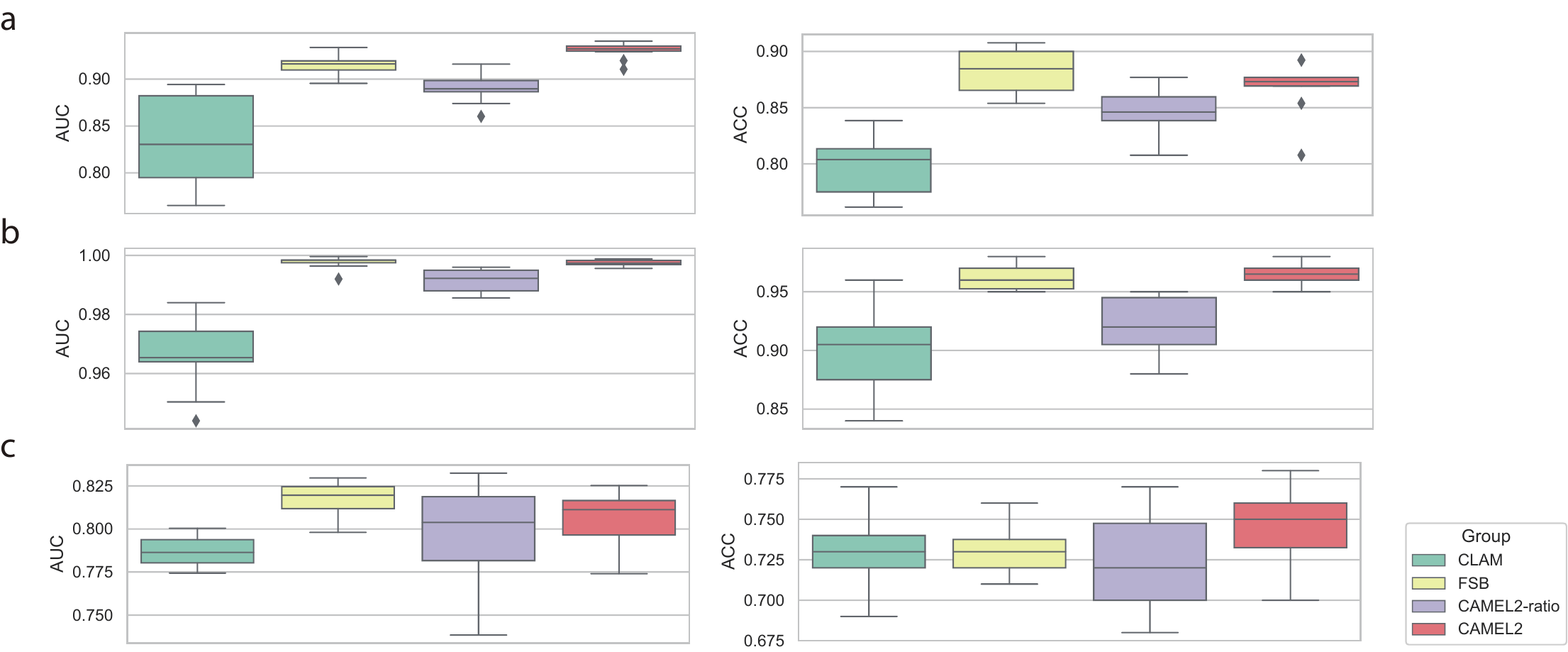}
    \caption{{\bf Slide-level aggregation results of different methods based on CLAM with the model type of "$clam_{mb}$".} {\bf a}, CAMELYON16. {\bf b}, In-house gastric cancer dataset. {\bf c}, In-house cervix cancer dataset.}
    \label{fig:slide}
\end{figure}

{\bf Slide-level aggregation.}
CLAM \cite{RN6} performs feature extraction for each instance using a model pretrained with ImageNet \cite{RN20}. Here, we replace CLAM's feature extraction model with our models and concatenate the 512-dimensional features from the last dense layer of ResNet-18 with the 2-dimensional softmax predictions as the extracted instance-level features. We then apply the same feature aggregation procedure as that in CLAM. The 10-fold Monte Carlo cross-validation results are presented in Figure \ref{fig:slide}.

With the help of some additional 5,120$\times$5,120 annotations, CAMEL2 surpasses CLAM in terms of the slide-level AUC and ACC. Moreover, the performance in terms of the slide-level prediction using the feature extraction model from CAMEL2 is comparable to that achieved by the FSB, which means that CAMEL2 can achieve comparable performance to that of the FSB measured by global slide-level prediction, and the relatively low patch-level sensitivity of CAMEL2 does not compromise the slide-level accuracy. It is worth noting that the instance-level features from CAMEL2 can also be integrated with other aggregation methods.

However, when replacing the feature extraction model with CAMEL2-slide, the accuracy of the slide-level prediction significantly decreases. For instance, the average AUC and ACC with the test set of the cervical cancer dataset drop to 0.74 and 0.70, respectively. This decrease in performance may be attributed to the low sensitivity of CAMEL2-slide. Therefore, for training a slide-level classifier using only slide-level annotations, directly enlarging the size of the bag in CAMEL2 to cover an entire WSI is inadequate.

{\bf ALK-rearrangement prediction.}
The in-house lung ALK dataset only contains slide-level annotations: positive for ALK+ slides and negative for ALK- slides. Previous studies \cite{RN26} showed that ALK+ lung adenocarcinomas have distinctive morphologic features, and it is difficult to distinguish an ALK mutation simply by histopathology imaging. Since the interpretability of CAMEL2 is reasonably good, in this study, we examine the capability of CAMEL2 to identify the regions related to ALK-rearrangement.

\begin{figure}
    \centering
    \includegraphics[width=0.9\textwidth]{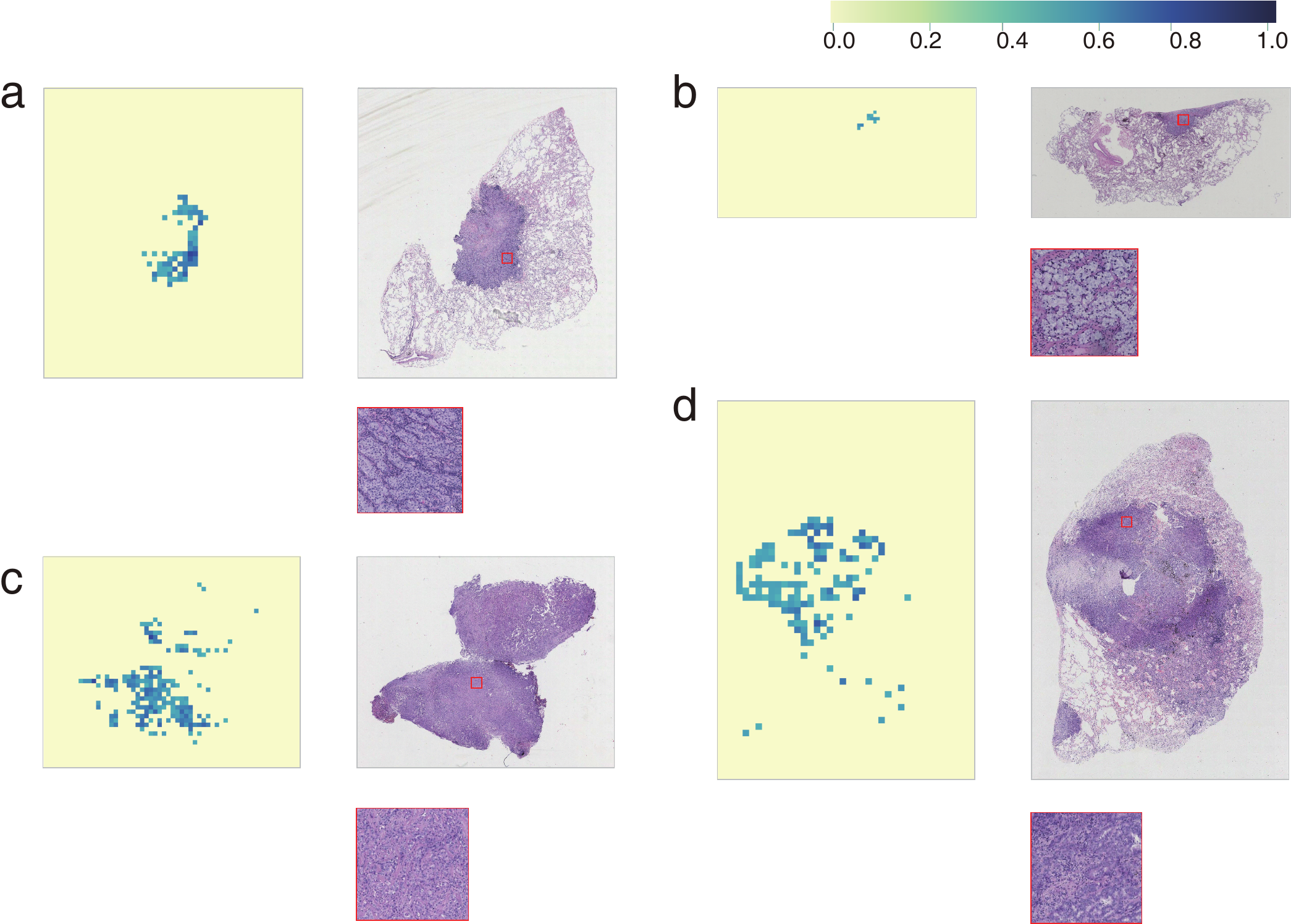}
    \caption{{\bf Results of CAMEL2 in identifying regions related to ALK-rearrangement.} {\bf a,b}, The examples on locating the regions with signet ring cells. {\bf c,d}, The examples on locating the regions with a solid growth pattern.}
    \label{fig:alk}
\end{figure}

Directly enlarging the size of the bag in CAMEL2 to cover an entire WSI may lead to a significant performance decrease; therefore, we first use an in-house lung cancer detection model to locate the instances that contain cancerous regions and use these instances to form a bag for each WSI. In this case, the noncancerous regions are removed. Note that both ALK+ slide and ALK- slides contains cancerous regions, and our object is to identify the areas related to the ALK-rearrangement from these cancerous regions.

The results indicate that our model successfully identifies regions with signet ring cells (Figures \ref{fig:alk}a and d), a histological feature commonly associated with ALK-rearranged adenocarcinoma \cite{RN27,RN28}. Additionally, the model detects regions with a solid growth pattern (Figures \ref{fig:alk}b and c), which has been previously linked to ALK-rearrangement in other studies \cite{RN26,RN29}.

\section*{Discussion}
In the field of histopathology image analysis, MIL-based weakly supervised methods \cite{RN3,RN6,RN7,RN10} play a crucial role, as WSIs are typically large in size. These methods generally involve two steps: feature extraction and feature aggregation. Most research focuses on the latter step by developing advanced aggregation algorithms to represent slide-level predictions from patch-level features obtained through feature extraction models pretrained with ImageNet \cite{RN20}. Due to the lack of informative annotations, although these methods are able to achieve reasonably good performance in terms of slide-level classification by using slide-level annotations exclusively, their performance in locating cancerous regions, which is vital for clinical applications, remains unsatisfactory. In contrast, our approach, CAMEL2, concentrates on the feature extraction step, aiming to train a feature extraction model with less labelling burden for the additionally introduced annotations but can achieve performance comparable to that of its fully supervised counterpart, thereby improving the interpretability of the weakly supervised learning method by effectively identifying cancerous regions.

By comparing the results between CAMEL2 and CLAM \cite{RN6}, we observe that methods focusing on feature aggregation, such as CLAM, excel in delivering slide-level predictions, as they primarily optimize loss functions for slide-level labels. On the other hand, CAMEL2, which emphasizes training a satisfactory instance-level classifier, excels in delivering instance-level predictions and offers better interpretability for clinical applications.

Traditional MIL-based methods typically select only one instance per bag, resulting in wasted information from the remaining instances. In CAMEL \cite{RN8}, we find that this will also cause a shift in the decision boundary towards the positive direction, resulting in low sensitivity and high specificity. To address this issue, we select two instances for each negative bag: one with the highest positive probability and the other with the highest negative probability. However, substantial information waste still exists, particularly when the bag size increases. Thus, in CAMEL2, we maximize information utilization by introducing the threshold of the cancerous ratio. We define positive bags as those with a cancerous ratio exceeding 20\% during the annotation process. In this case, we can effectively utilize 20\% of the information in each positive bag and all the information in each negative bag during training. Meanwhile, the additional required labelling workload is relatively small. For instance, for training CAMEL2 with CAMELYON16, only an additional 322 5,120$\times$5,120 positive images with cancerous ratio higher than 20\% from 110 positive WSIs are needed.

To verify the effectiveness of CAMEL2, we examine its capability in locating the pseudo target with a commonly used natural image dataset, MNIST \cite{RN24}, and in identifying the cancerous instances with three histopathology image datasets (CAMELYON16 \cite{RN23}, an in-house gastric cancer dataset \cite{RN1}, and an in-house cervical cancer dataset). The results with MNIST demonstrate that CAMEL2 successfully identifies the target instances in a weakly supervised manner for the bags with different proportions of target, showcasing the feasibility of our straightforward strategy. Furthermore, with the histopathology image datasets, CAMEL2 achieves comparable instance-level classification performance to that of the FSB with the help of 5,120$\times$5,120 image-level binary annotations, which are easy annotate. In addition, with the ALK dataset, by introducing the information of the predicted cancerous regions from other models, CAMEL2 successfully locates some regions that may be related to the ALK rearrangements, indicating the potential of integrating CAMEL2 with other models to achieve better interpretability.

In CAMEL2, we initially estimate the significance ratio for a specific dataset and use it to determine the number of instances included in positive bags. To showcase the performance when the correct significance ratio is applied, we introduce CAMEL2-ratio as a benchmark. The results indicate that CAMEL2, in terms of certain metrics, has outperformed CAMEL2-ratio. This suggests that even a reasonably estimated ratio can produce satisfactory outcomes.

Currently, the binary labeling of 5,120$\times$5,120 images requires pathologists' expertise. Our results with the histopathology datasets reveal that directly enlarging the bag size in CAMEL2 from 5,120$\times$5,120 to an entire WSI (CAMEL2-slide) leads to a significant performance decrease. This drop may be attributed to the confusion arising from the substantial increase in noncancerous regions within positive bags. Thus, in the future, we plan to explore a method for labelling the 5,120$\times$5,120 images using slide-level annotations exclusively. The main challenge lies in dealing with the extremely large dimensions of each instance (i.e., 5,120$\times$5,120 images). Fortunately, the size of each bag is typically small. For example, for the illustrative WSI with 44,800$\times$43,008 pixels shown in Figure \ref{fig:framework}a, only 33 instances remain in its bag after filtering out the background.

\section*{Methods}

{\bf Ethical approval.}
The study was approved by the institutional review board of each participating hospital (Medical Ethics Committee, Chinese PLA General Hospital; Ethics Committee of China-Japan Friendship Hospital). The informed consents were waived by the institutional review boards since the reports were anonymized.

{\bf Datasets.}
We evaluate CAMEL2 with one natural image dataset MNIST \cite{RN24}; three histopathology image datasets: CAMELYON16 \cite{RN23}, an in-house gastric cancer dataset \cite{RN1} and an in-house cervical cancer dataset; and an in-house ALK-rearranged lung adenocarcinomas dataset.

MNIST \cite{RN24} is a commonly used dataset for evaluating computer version algorithms, consisting of 60,000 examples of handwritten digits ("0" to "9") in the training set and 10,000 examples in the test set.

CAMELYON16 \cite{RN23} is a public dataset with 399 hematoxylin-eosin (H\&E) stained WSIs of lymph node sections. The official training set of CAMELYON16 contains 270 WSIs, including 110 positive and 160 negative slides. In our research, we treat the 5,120$\times$5,120 patches at 20$\times$ magnification in the WSIs as image-level data. We split WSIs into non-overlapping 5,120$\times$5,120 image-level data and filter out the background using the Otsu's method \cite{RN30}. For training CAMEL2-ratio, we use 727 positive images with ratio constraints from positive WSIs and 6,651 negative images without any cancerous regions. For training CAMEL2, we select 322 positive images with a cancerous ratio higher than 20\% from positive WSIs. The official test set of CAMELYON16 consists of 129 WSIs, including 49 positive and 80 negative slides. To evaluate the instance-level (320$\times$320) classification accuracy, we split these images into 320$\times$230 instances at 20$\times$ magnification and filter out the background. An instance is labelled as positive if it contains any cancerous pixels; otherwise, it is labelled as negative.

We use a subset of gastric cancer dataset collected by Song et al. \cite{RN1} in this study. This dataset includes 420 WSIs (170 positive and 250 negative images). We randomly select 120 positive and 200 negative images for the training set and 50 positive and 50 negative slides for the test set. For training CAMEL2-ratio, 1,031 positive 5,120$\times$5,120 images with specific ratio constraint from positive WSIs and 8,302 negative images are used. For training CAMEL2, 758 positive images with a cancerous ratio higher than 20\% from positive WSIs are adopted.

We have also collected an in-house cervical cancer dataset with 1,156 WSIs, including 444 positive and 712 negative images. We randomly select 394 positive and 662 negative images for the training set and 50 positive and 50 negative images for the test set. Similar to the preprocessing step for CAMELYON16, for training CAMEL2-ratio, 669 positive 5,120$\times$5,120 images with specific ratio constraints from positive WSIs and 4,023 negative images are used. For training CAMEL2, 291 positive images with a cancerous ratio higher than 20\% from positive WSIs are adopted.

Constitutive activation of anaplastic lymphoma receptor tyrosine kinase (ALK) caused by chromosomal rearrangement defines a category of lung adenocarcinomas that may be amenable to targeted therapy with the ALK inhibitor crizotinib \cite{RN31}. In this study, we have constructed an in-house ALK dataset that contains 553 WSIs (165 positive ALK+ and 388 negative ALK- images).

\textbf{Metrics.}
To evaluate the instance-level binary classification performance at 320$\times$320 resolution, we use metrics such as sensitivity, specificity, F1 score, and the AUC. To evaluate the slide-level binary classification performance, we utilize metrics such as classification accuracy and AUC.

\textbf{Framework of CAMEL2.}
In CAMEL2 and CAMEL2-ratio, we treat a 5,120$\times$5,120 image-level patch as a bag and split it into 256 instances of size 320$\times$320. These instances are then fed into an instance classifier (ResNet-18 \cite{RN25}). As shown in Figure \ref{fig:framework}b, during the forwards pass, we calculate the binary classification probability for each instance. During backpropagation, for a positive bag, we sort the instances based on their positive probabilities and select the top $t$\% instances with the highest positive probability. The selected instances are assigned positive labels. For a negative bag, we use all instances within it, and negative labels are assigned. Finally, we update the classifier's parameters using the cross-entropy loss calculated by the selected instances.

We also examine the performance of CAMEL2 using the entire WSI as a bag (CAMEL2-slide). In this case, we only utilize slide-level annotations. Similar to CAMEL2, we split a WSI into instances of size 320$\times$320. Due to GPU memory limitations, we randomly select 512 instances for each WSI. If the number of instances is less than 512, we use all instances. We set the threshold for a positive bag to 10\% since the WSI size is much larger than 5120x5120. This means that only instances with the top 10\% positive probability in each positive WSI are considered positive and included in the training process. In short, the threshold indicates the ratio of utilized instances in each positive bag.

Inspired by the retraining  step in CAMEL \cite{RN8}, we employ a similar approach in CAMEL2. After training the classifier, we use it to classify all instances in each positive bag, and the predicted positive instances are saved. Then, we retrain a new classifier from scratch. This time, during training, we only select instances from the previously saved instances for each positive bag. If the ratio of the saved positive instances in a bag is greater than 20\%, we choose the top 20\% instances with the highest predicted positive probability. Otherwise, we include all instances.

\textbf{Training details.}
CAMEL2 is implemented in TensorFlow \cite{RN32} and trained using four NVIDIA Tesla V100 GPUs. The batch size for each model is set to 16, with 2 positive bags and 2 negative bags on each GPU. We initialize the weights of the classifier using the Glorot uniform initializer. The Adam optimizer \cite{RN33} is utilized with an initial learning rate of 0.001. We train for 20 epochs, and the learning rate is halved after every 5 epochs. In CAMEL2, we use 5 models, which are saved after the 12th, 14th, 16th, 18th, and 20th training epochs, to form an ensemble instance classifier. The final output is obtained by calculating the mean of their predicted positive probabilities.

\begin{addendum}
 \item[Acknowledgements] This work is supported by the National High Level Hospital Clinical Research Funding of China (No. 2022-NHLHCRF-LX-01-0206), National Natural Science Foundation of China (No. 32300535), National Key Research and Development Program of China (No. 2021YFF1200400), 2023 Science and Technology Projects of Qinghai Province, China (Basic Research Program, No. 2023-ZJ-732).
 \item[Author Contributions Statement] G.X. and S.W. proposed the research, L.Z., Z.S. and A.L. performed the WSI acquisition, G.X., X.C. and L.W. conducted the experiment, G.X. wrote the deep learning code and performed the experiment, G.X., L.Z., X.C., T.W., Z.L., Z.Z., W.B., Z.S. and S.W. wrote the manuscript, D.W., D.Z., A.L., H.S. and J.M. reviewed the manuscript.
 \item[Competing Interests] Shuhao Wang is the co-founder and chief technology officer (CTO) of Thorough Future. Lang Wang and Zewen Zhang are algorithm researchers of Thorough Future. All remaining authors have declared no conflicts of interest.
 \item[Data Availability] The predictions for the CAMELYON16 testset is available at \url{https://github.com/ThoroughFuture/CAMEL2}. The data that support the findings of this study are available on request from the corresponding authors (S.W., D.Z. and J.M.).
\item[Code Availability] The code of CAMEL2 is available at \url{https://github.com/ThoroughFuture/CAMEL2}.
 \item[Correspondence] Correspondence and requests for materials
should be addressed to S.W.~(email: \mbox{to@shuhao.wang}), D.Z.~(email: zhongdingrong@sina.com), and J.M.~(email: \mbox{jpma@fudan.edu.cn}).
\end{addendum}

\bibliography{ref}

\clearpage

\setcounter{page}{1}
\setcounter{figure}{0}
\setcounter{table}{0}
\renewcommand{\thetable}{S\arabic{table}}
\renewcommand{\thefigure}{S\arabic{figure}}

\begin{centering}

~\\
~\\

{\bf \Large CAMEL2: Enhancing weakly supervised learning for histopathology images by incorporating the significance ratio}

{\large Xu et al.}

\end{centering}

\begin{figure}[H]
    \centering
    \includegraphics[width=0.8\textwidth]{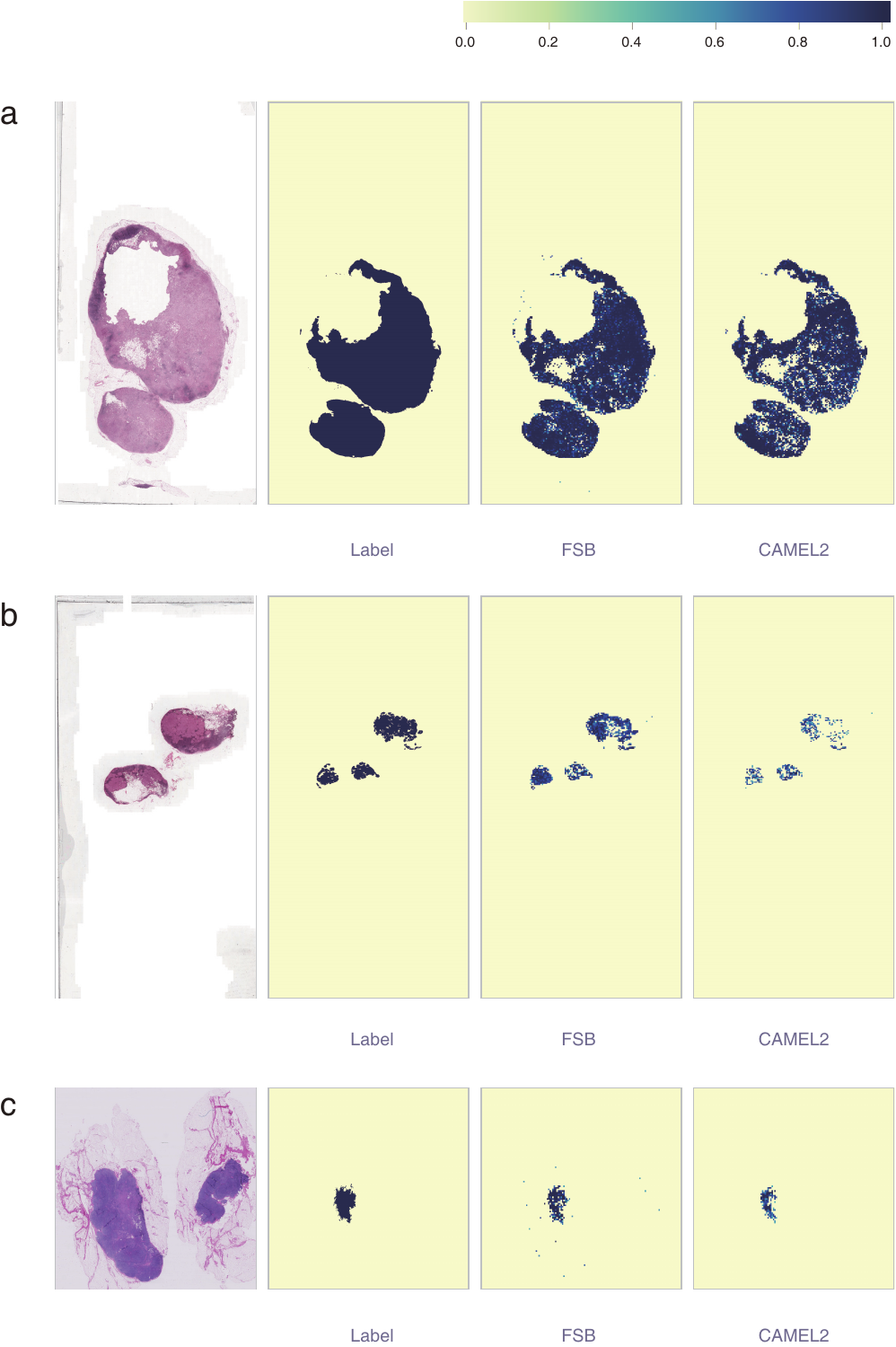}
    \caption{{\bf Examples of low sensitivity cases from CAMEL2 with the test set of CAMELYON16.}}
    \label{fig:lsen}
\end{figure}

\begin{figure}[H]
    \centering
    \includegraphics[width=0.9\textwidth]{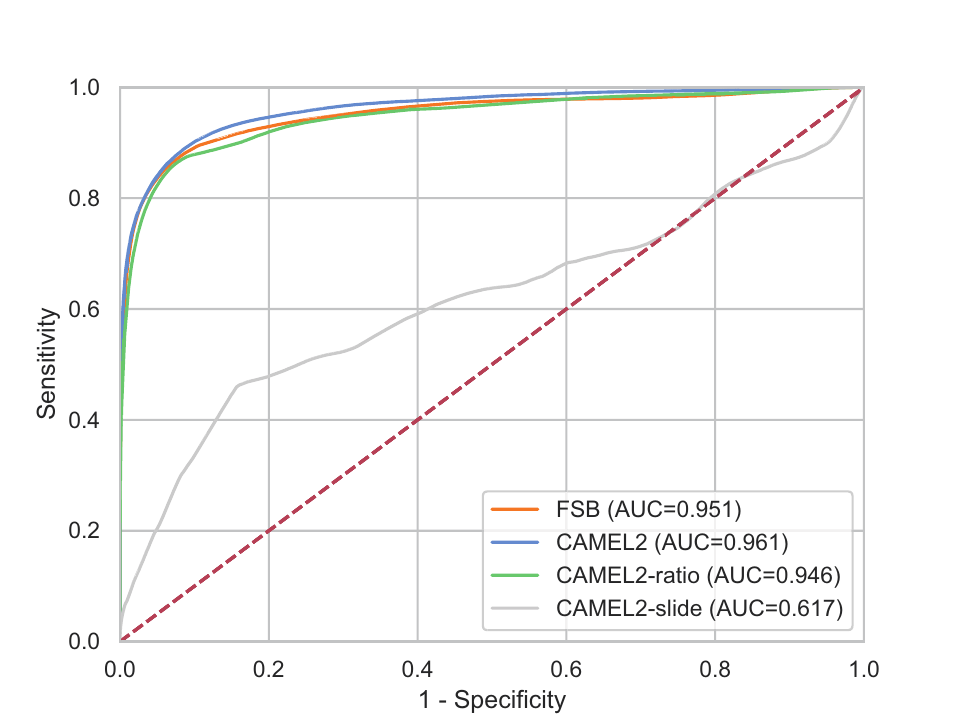}
    \caption{{\bf The ROC curve of the 320$\times$320 patch-level binary classification results on the test set of in-house gastric cancer dataset.}}
    \label{fig:roc_gastric}
\end{figure}

\begin{figure}[H]
    \centering
    \includegraphics[width=0.9\textwidth]{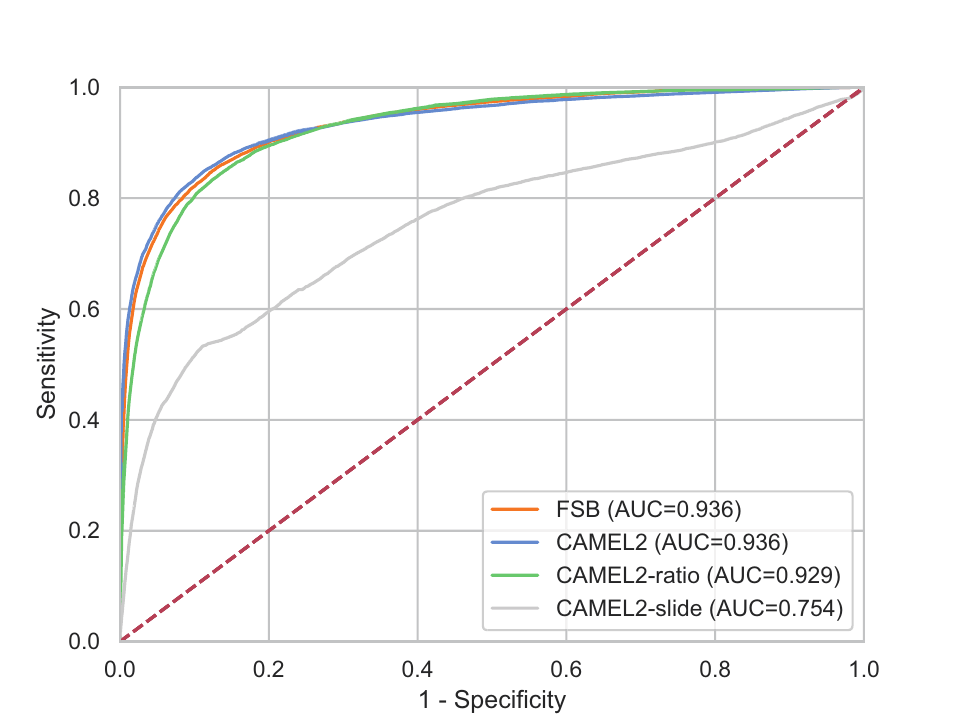}
    \caption{{\bf The ROC curve of the 320$\times$320 patch-level binary classification results on the test set of in-house cervix cancer dataset.}}
    \label{fig:roc_cervix}
\end{figure}

\begin{figure}[H]
    \centering
    \includegraphics[width=1.0\textwidth]{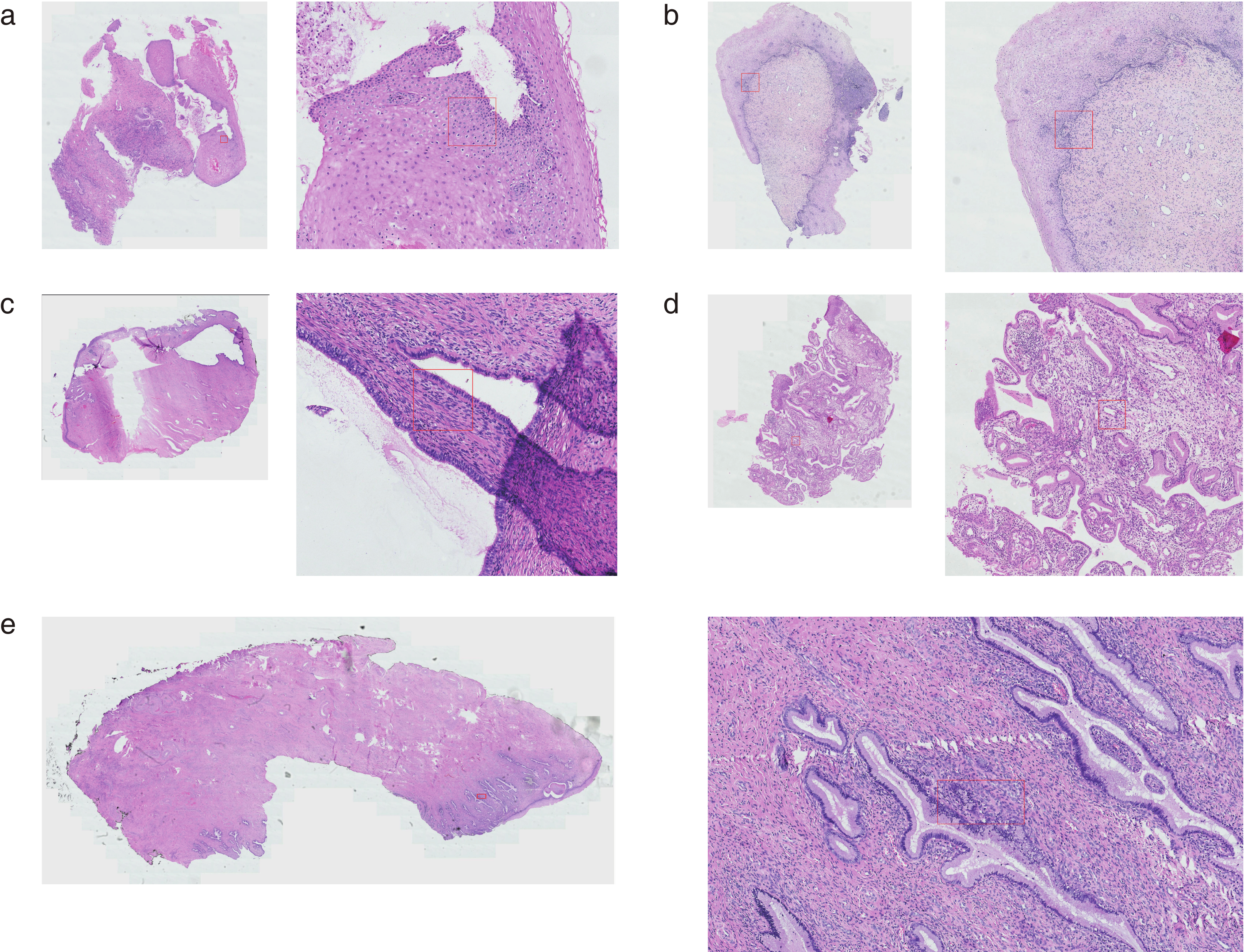}
    \caption{{\bf Typical false positive predictions of CAMEL on the in-house cervix dataset.} {\bf a}, Low-grade cervical squamous intraepithelial lesion. {\bf b}, Low-grade cervical squamous intraepithelial lesion. {\bf c}, Normal gland. {\bf d}, Blood vessel with an irregular shape. {\bf e}, Normal gland.}
    \label{fig:cervix_fp}
\end{figure}

\begin{figure}[H]
    \centering
    \includegraphics[width=1.0\textwidth]{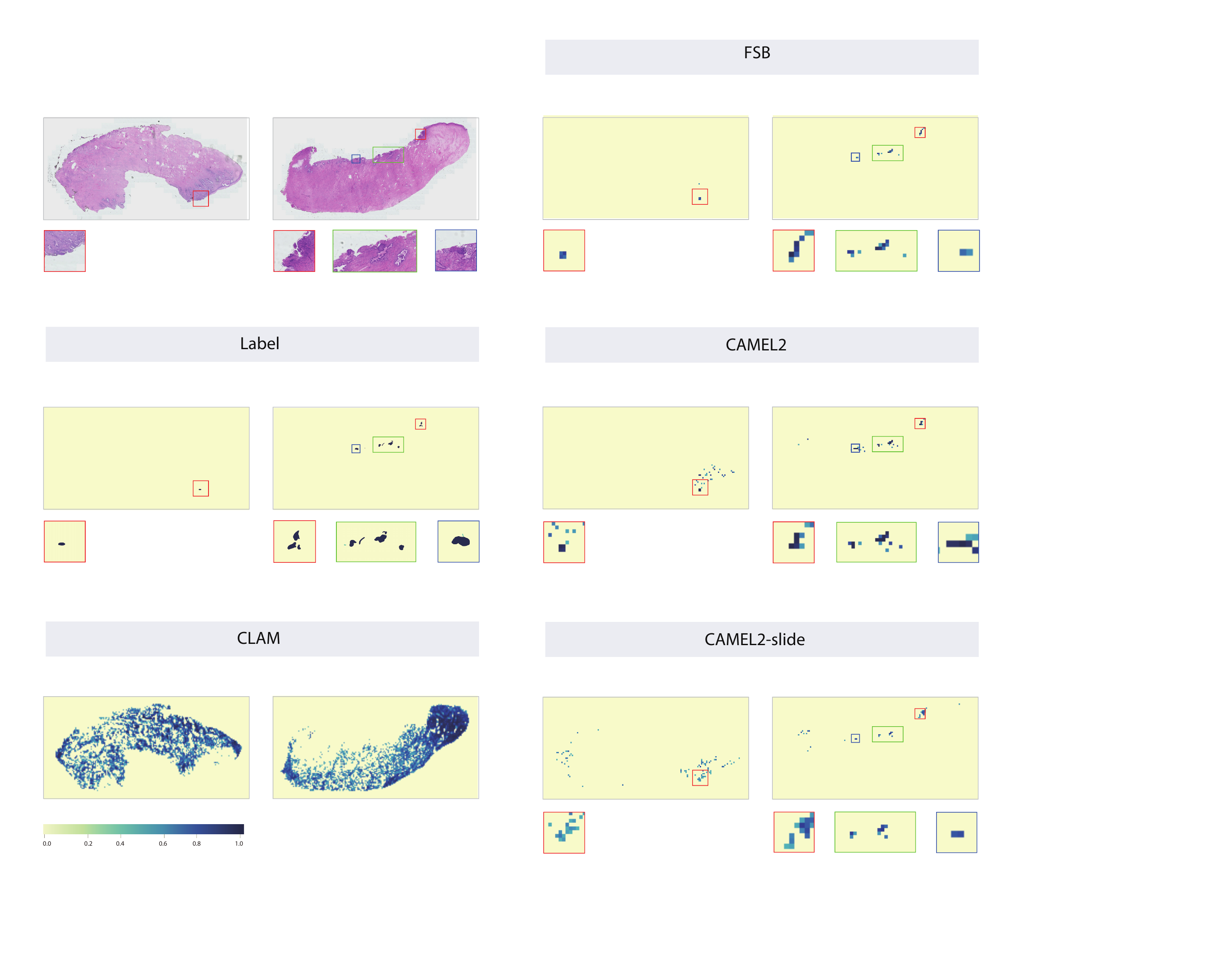}
    \caption{{\bf Performance of different methods using test set WSIs of the cervical cancer dataset with a relatively small ratio of cancerous regions.}}
    \label{fig:cervix_region}
\end{figure}

\begin{table}
  \centering
  \caption{The average time taken by three pathologists in preparing annotations for different methods on 5 positive WSIs from CAMELYON16.}
  \label{tab:s1r}
    \begin{tabular}{lllll}
    \\ \hline
          & CAMEL2-slide & CAMEL2 & CAMEL2-ratio    & Fully supervised \\
    \hline
    Time (min)   & 0.0 & 12.2 & 19.6 & 45.8 \\
    \hline
    \end{tabular}
\end{table}

\begin{table}
  \centering
  \caption{320x320 patch-level binary classification results of different methods with the official test set of CAMELYON16.}
    \begin{tabular}{lllll}
    \\ \toprule
          & Sensitivity & Specificity & F1    & AUC \\
    \midrule
    \multicolumn{5}{l}{320$\times$320 instance-level annotation (20$\times$ magnification)} \\
    \midrule
    FSB   & 0.838 & 0.996 & 0.869 & 0.972 \\
    \midrule
    \multicolumn{5}{l}{1,280$\times$1,280 image-level annotation (20$\times$ magnification)} \\
    \midrule
    MIL   & 0.653 & 1     & 0.788 & 0.968 \\
    CAMEL   & 0.748 & 0.999     & 0.843 & 0.971 \\
    CAMEL2   & 0.755 & 0.998     & 0.833 & 0.975 \\
    CAMEL2-ratio & 0.826 & 0.997 & 0.873 & 0.966 \\
    \midrule
    \multicolumn{5}{l}{5,120$\times$5,120 image-level annotation (20$\times$ magnification)} \\
    \midrule
    MIL   & 0.352 & 1     & 0.52  & 0.949 \\
    CAMEL   & 0.143 & 1     & 0.251  & 0.866 \\
    CAMEL2-ratio & 0.747 & 0.998 & 0.831 & 0.979 \\
    CAMEL2-ratio wo/retrain & 0.755 & 0.997 & 0.829 & 0.974 \\
    CAMEL2 & 0.737 & 1     & 0.844 & 0.97 \\
    CAMEL2 wo/retrain & 0.735 & 0.999 & 0.839 & 0.969 \\
    CAMEL2 ($t$ = 50\%) & 0.653 & 0.999 & 0.782 & 0.959 \\
    CAMEL2 ($wt$ = 25\%) & 0.748 & 0.998 & 0.833 & 0.975 \\
    CAMEL2 ($wt$ = 35\%) & 0.774 & 0.994 & 0.806 & 0.957 \\
    \midrule
    \multicolumn{5}{l}{Slide-level annotation (20$\times$ magnification)} \\
    \midrule
    CAMEL2-slide & 0.383 & 0.995  & 0.508   & 0.845 \\
    \bottomrule
    \end{tabular}
  \label{tab:s1}
\end{table}

\begin{table}
  \centering
  \caption{320x320 patch-level binary classification results of different methods using the test set of the in-house gastric cancer dataset.}
    \begin{tabular}{lllll}
    \\ \toprule
          & Sensitivity & Specificity & F1    & AUC \\
    \midrule
    \multicolumn{5}{l}{320$\times$320 instance-level annotation (20$\times$ magnification)} \\
    \midrule
    FSB   & 0.852 & 0.94  & 0.755 & 0.951 \\
    \midrule
    \multicolumn{5}{l}{5,120$\times$5,120 image-level annotation (20$\times$ magnification)} \\
    \midrule
    CAMEL2-ratio & 0.861 & 0.926 & 0.729 & 0.946 \\
    CAMEL2 & 0.636 & 0.995 & 0.76  & 0.961 \\
    \midrule
    \multicolumn{5}{l}{Slide-level annotation (20$\times$ magnification)} \\
    \midrule
    CAMEL2-slide & 0.159 & 0.965 & 0.228 & 0.617 \\
    \bottomrule
    \end{tabular}
  \label{tab:s2}
\end{table}

\begin{table}
  \centering
  \caption{320x320 patch-level binary classification results of different methods with the test set of the in-house cervical cancer dataset.}
  \label{tab:s3}
    \begin{tabular}{lllll}
    \\ \hline
          & Sensitivity & Specificity & F1    & AUC \\
    \hline
    \multicolumn{5}{l}{320$\times$320 instance-level annotation (20$\times$ magnification)} \\
    \hline
    FSB   & 0.578 & 0.985 & 0.653 & 0.936 \\
    \hline
    \multicolumn{5}{l}{5,120$\times$5,120 image-level annotation (20$\times$ magnification)} \\
    \hline
    CAMEL2-ratio & 0.741 & 0.931 & 0.563 & 0.929 \\
    CAMEL2 & 0.695 & 0.97  & 0.669 & 0.936 \\
    \hline
    \multicolumn{5}{l}{Slide-level annotation (20$\times$ magnification)} \\
    \hline
    CAMEL2-slide & 0.29  & 0.975 & 0.36  & 0.754 \\
    \hline
    \end{tabular}
\end{table}

\end{document}